%% file: root.tex
\documentclass[a4paper, 10 pt, conference]{ieeeconf}      

\IEEEoverridecommandlockouts                              

\usepackage{graphics} 
\usepackage{epsfig} 
\usepackage{mathptmx} 
\usepackage{times} 
\usepackage{amsmath} 
\usepackage{amssymb}  
\usepackage{soul}
\usepackage{lipsum}
\usepackage[binary-units=true,per-mode=symbol]{siunitx}
\usepackage{epstopdf}
\usepackage[space]{cite}
\usepackage[nolist]{acronym} 
\usepackage{hyperref} 
\usepackage[]{algorithm,algpseudocode,placeins}
\usepackage{caption}
\usepackage{subcaption}
\usepackage[a4paper, left=0.675in, right=0.675in, bottom=1.2in, top=0.83in]{geometry}

\newcommand*\numcircledmod[1]{\raisebox{.5pt}{\textcircled{\raisebox{-.9pt} {#1}}}}

\title{\LARGE \bf
Steering Action-aware Adaptive Cruise Control\\for Teleoperated Driving
}

\author{Andreas Schimpe, Domagoj Majstorovic and Frank Diermeyer
\thanks{The authors are with the Institute for Automotive Technology at the Technical University of Munich (TUM), 85748 Garching bei M\"unchen, Germany. {\tt\small andreas.schimpe@tum.de}}%
}

\begin{document}
\maketitle
\thispagestyle{empty}
\pagestyle{empty}

\input{acronyms.tex}
\begin{abstract}
In this paper, a steering action-aware Adaptive Cruise Control (ACC) approach for teleoperated road vehicles is proposed. In order to keep the vehicle in a safe state, the ACC approach can override the human operator's velocity control commands. The safe state is defined as a state from which the vehicle can be stopped safely, no matter which steering actions are applied by the operator. This is achieved by first sampling various potential future trajectories. In a second stage, assuming the trajectory with the highest risk, a safe and comfortable velocity profile is optimized. This yields a safe velocity control command for the vehicle. In simulations, the characteristics of the approach are compared to a Model Predictive Control-based approach that is capable of overriding both, the commanded steering angle as well as the velocity. Furthermore, in teleoperation experiments with a 1:10-scale vehicle testbed, it is demonstrated that the proposed ACC approach keeps the vehicle safe, even if the control commands from the operator would have resulted in a collision.
\end{abstract}
\section{\uppercase{Introduction}}
\label{sec:introduction}
The technology of road vehicle teleoperation, i.e., \ac{tod} holds the promise to achieve truly driverless \acp{av}~\cite{Georg2018keytech, Neumeier2018way2ad}. It is assumed that the \ac{av} is capable of fully \ac{ad}. However, at some point, the function may reach a constraint of its operational design domain, e.g., an intersection that is controlled by a policeman. Resolving such a traffic situation may not yet be covered by the \ac{ad} function. Instead, the vehicle comes to a stop and requests support from a remote \ac{vcoc}. From there, a human operator connects to the vehicle via mobile networks. Transmission of sensor data, in particular vehicle state information  and video streams, from the vehicle to the \ac{vcoc} is initiated. There, the data are displayed to the operator, giving him an understanding of the vehicle surroundings. Based on this, the operator is able to generate control commands, which are transmitted back to the vehicle for execution. \\
\ac{tod} comes with challenges. Data, as perceived by an operator, are subject to latency due to processing and the transmission via a mobile network~\cite{Georg2020latency}, which also yields the risk of an unstable connection~\cite{Hoffmann2021safetyAssessment}. Furthermore, due to various reasons, the \ac{sa} of an operator in the \ac{vcoc} may be lower than the \ac{sa} of a driver in the vehicle. Frequently described issues are the ``out-of-the-loop-syndrome,'' having to deal with an unfamiliar traffic situation at short notice, or an operator's missing sense of embodiment, i.e., an adequate feeling for the actual motion of the controlled vehicle~\cite{Mutzenich2021situationAwareness}. 
\subsection{Vehicle Teleoperation Concepts}
\label{sec:teleopConcepts}
Remote assistance or control concepts can be designed in different ways in order to resolve the various fail cases of a complex \ac{ad} function. For instance, the concept in~\cite{Feiler2021remotePercMod} aims at enabling remote assistance for tasks related to perception. Interactive path planning approaches~\cite{Hosseini2014interactivePathPlanning, Schitz2021interactivePathPlanning} assume a functional perception module and support the \ac{ad} function at the decision making layer. While the human operator takes over the legal responsibility for this decision, it is the responsibility of the \ac{ad} function to ensure the safety of the vehicle during execution. \\
At the lowest layer, remote assistance at the control layer is possible through direct teleoperation, i.e., an operator is controlling the vehicle at stabilization level. Through a conventional control interface, consisting of a steering wheel and pedals, the operator generates lateral and longitudinal motion control commands, which are sent to the vehicle actuators via the mobile network. This control concept makes minimal to no assumptions about the operation of \ac{ad} function modules. However, especially due to the previously described challenges of \ac{tod}, it also yields the greatest workload for the operator, who essentially takes over all aforementioned tasks of the \ac{ad} function, namely correct perception, legal decision making as well as the safe motion of the vehicle. In the presented work, shared control is seen as a vehicle teleoperation concept that keeps the convenience of a conventional control interface, but supports the operator in ensuring the safety of the vehicle. 
\subsection{Uncoupled Shared Control for Vehicle Teleoperation}
\label{sec:sharedCtrl}
Shared control is a case of human-machine interaction. In a shared control system, a human and an automation component are collaborating in performing a task, e.g., controlling a vehicle. In this context, as introduced in~\cite{Marcano2020review}, it can be distinguished between coupled and uncoupled shared control designs. In the coupled case, the partners, human and machine, are acting on mechanically coupled actuators. Authority is held by the partner applying the greater force. In contrast to this, uncoupled shared control is applicable in a drive-by-wire system. The automation is receiving the control actions from the human as a reference and can override if deemed necessary, e.g., in order to avoid an imminent collision of the controlled vehicle. Being the focus of the approach proposed in this paper, this capability makes uncoupled shared control in particular of great interest for the use case of vehicle teleoperation~\cite{Majstorovic2022rccs}.
\newpage
\subsection{Related Work}
\label{sec:relatedwork}
Various uncoupled shared control designs have been proposed. Several approaches assist the human driver or operator in ensuring safe lateral motion control of teleoperated vehicles~\cite{Anderson2010optimalCtrlFramework, Schimpe20steerwithme} or experimental drive-by-wire vehicles operated at high speed~\cite{Erlien2016sharedCtrlEnvelopes}. It is obvious that those approaches can no longer ensure vehicle safety if a collision can only be avoided through braking. In consequence, multiple approaches address this issue, given the capability of also overriding the human's longitudinal control command, i.e., desired velocity or acceleration. This has proven to effectively improve vehicle safety in environments with static~\cite{Storms2017sharedControl4oa} and dynamic obstacles~\cite{Schwarting2018parallelAutonomyWdynMdl, Saparia21ass4tod}. \\
Throughout the years, cruise control, which only decouples longitudinal control from the driver, has been designed for different tasks and purposes. For instance, predictive cruise control approaches aim at improving fuel economy through the use of upcoming traffic signal information~\cite{Asadi2011acc} or high-definition maps~\cite{Chu2018acc}. \ac{acc} designs have the objective to keep the controlled vehicle safe, while it is performing lane keeping and car following. Some designs specifically address safety in the case of unexpected behavior of surrounding vehicles, e.g., cut-in situations~\cite{Althoff2021acc} or full braking of the preceding vehicle~\cite{Magdici2017acc}. Furthermore, cooperative \ac{acc}, the controlled vehicle being connected to the preceding vehicle, has been shown to enable smaller safety distances and improve traffic-flow stability~\cite{VanArem2006acc}. The \ac{acc} approach in~\cite{Schitz2021acc}, developed for \ac{tod} tasks, considers urban scenarios involving cross traffic. \\
Classical \ac{acc} approaches assume that the controlled vehicle is performing lane keeping. Unforeseen lateral maneuvers, e.g., sudden lane changes or turns in urban scenarios, are not considered and thus safety would not be ensured. For the challenging task of \ac{tod}, steering actions from the human operator may not be safe at all times. 
\subsection{Contributions}
\label{sec:contribs}
In this paper, a steering action-aware \ac{acc} approach for \ac{tod} is proposed. As an active safety system, the approach can override the human operator's velocity control commands. As visualized in Fig.~\ref{fig:architecture}, it consists of two stages. First, various future steering actions are evaluated. Second, a safe velocity profile is optimized. In simulations, the characteristics of the \ac{acc} approach are compared to a baseline shared controller, capable of also overriding the commanded steering angle. Furthermore, the approach is validated in experiments with an actual operator, teleoperating a 1:10-scale vehicle testbed.  
\section{\uppercase{Approach}}
\label{sec:approach}
\begin{figure}[!t]
    \includegraphics[width=\linewidth]{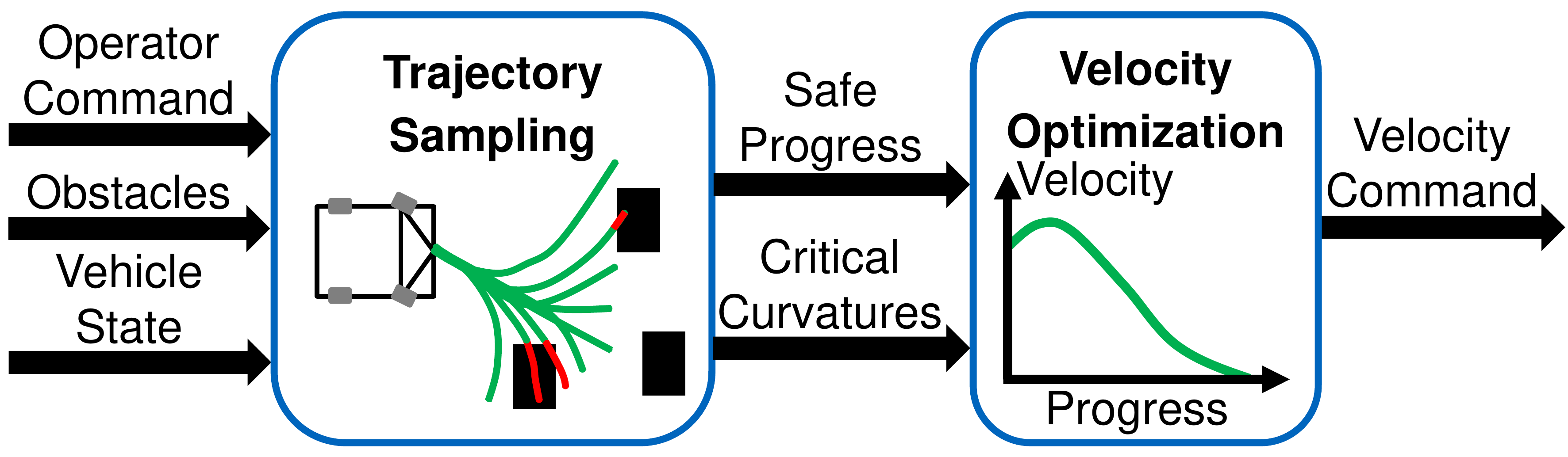}
    \caption{Stages of Steering Action-aware \ac{acc} Approach.}
    \label{fig:architecture}
\end{figure}

The procedure, corresponding to the stages of the steering action-aware \ac{acc} approach, visualized in Fig.~\ref{fig:architecture}, is given in~Alg.~\ref{alg:cmpVelCmd}. From the current vehicle state with the steering angle~$\delta_{\textrm{curr}}$ and velocity~$v_{\textrm{curr}}$, the~\texttt{TreePlanner} samples various future steering actions in line~2. In line 3, the generated trajectories are checked for collisions with the list of obstacles~$O$. Thereby, the collision-free, i.e., the safe progress in meters along each trajectory is computed. The minimal safe progress from all trajectories, i.e., the global safe progress~$s_{\textrm{safe}}$ is taken in line~4. Also from the~\texttt{TreePlanner}, the critical curvature profile~$\kappa_{\textrm{crit}}^N$ is taken in line~5. Here and throughout the remainder of this section, superscript~$N$ denotes the course of the corresponding variable over the planning horizon, discretized in~$N$ steps. In line~6, from the current velocity and acceleration~$a_{\textrm{curr}}$ of the vehicle, the~\texttt{VelocityOptimizer} solves for a velocity profile, taking into account the desired velocity of the human operator~$v_{\textrm{des}}$, the global safe progress and the critical curvature profile. From this profile, the entry at time instant~$1$ is returned from the algorithm in line~7 as the velocity command to be executed by the vehicle. \\
In the remainder of this section, more details on the above steps are provided. For brevity, they are summarized as~\textit{A.~Trajectory Sampling}~(lines~2 to~5), yielding the global safe progress and the critical curvature profile, and~\textit{B.~Velocity Optimization}~(lines~6 and~7), computing the safe velocity control command for the vehicle. 
\FloatBarrier
\begin{algorithm}[!t]
\small
\caption{Compute Adaptive Cruise Control Command}
\label{alg:cmpVelCmd}
\begin{algorithmic}[1]
\Procedure{$\mathbf{ComputeCommand}$}{$\delta_{\textrm{curr}}$, $v_{\textrm{curr}}$, $a_{\textrm{curr}}$, $v_{\textrm{des}}$, $O$}
    \State TreePlanner.GenerateTrajectoriesFrom($\delta_{\textrm{curr}}$, $v_{\textrm{curr}}$)
    \State TreePlanner.CheckForCollisionsWith($O$)
    \State $s_{\textrm{safe}} \gets$ TreePlanner.GetGlobalSafeProgress()
    \State $\kappa_{\textrm{crit}}^N \gets$ TreePlanner.GetCriticalCurvatureProfile()
    \State $v_{\textrm{opt}}^N \gets$ VelocityOptimizer.solve($v_{\textrm{curr}}$, $a_{\textrm{curr}}$, $v_{\textrm{des}}$, $s_{\textrm{safe}}$, $\kappa_{\textrm{crit}}^N$)
    \State \textbf{return} $v_{\textrm{opt}}(1)$
\EndProcedure
\end{algorithmic}
\end{algorithm}%
\subsection{Trajectory Sampling}
\label{sec:trajSampling}
At each sampling instant of the controller, a trajectory tree, a set of trajectories of time horizon~$T_H$, discretized in~$N$ steps by the sampling time~$t_s$, is generated. Assuming only moderate speeds during vehicle teleoperation, the well-known kinematic bicycle model~\cite{Rajamani2012vehicleDynamics} is used in order to plan the trajectories. For this, the system states are~$z = [x, y, \theta, \delta, v ]^T$, with the \ac{com} position of the vehicle in~$x$ and $y$, the heading~$\theta$, the steering angle~$\delta$ and the velocity~$v$. The inputs to the system are~$u = [\dot{\delta}, a]^T$, with the steering angle rate~$\dot{\delta}$ and the acceleration~$a$. 
%
\subsubsection{Trajectory Generation}
As pointed out earlier, the \ac{acc} evaluates various future steering actions commanded by the human operator within the planning horizon. The trajectory states are computed using the Forward Euler method, given by~$z_{n+1} = z_n + t_s \, \dot{z}_n$ for $n = 0, 1, ... \, N-1$. \\
The trajectories of the tree are planned by varying the steering angle rate in~$M$ steps within a steering angle rate range~$[-\dot{\delta}_{\textrm{max}} , \dot{\delta}_{\textrm{max}}]$, with an assumed maximum steering angle rate~$\dot{\delta}_{\textrm{max}}$. All trajectories plan to brake the vehicle to a standstill by applying a constant deceleration~$a_{\textrm{stop}} = -v_{\textrm{curr}}/T_H$. In summary, the~$M$ trajectories of the tree are planned by applying constant control inputs~$u_m = [ \dot{\delta}_m , a_{\textrm{stop}} ]^T$ with $\dot{\delta}_m = -\dot{\delta}_{\textrm{max}} + 2 \, \dot{\delta}_{\textrm{max}} (m-1) \, / (M-1)$ for $m = 1, 2, ... \, M$. \\
In Fig.~\ref{fig:accProcedureViz}, with parameters from a passenger vehicle, a snapshot of generated trajectories is visualized. In Fig.~\ref{fig:accTrajTree}, the trajectories are shown in the $xy$ plane. Fig.~\ref{fig:accStatesOverProgress} plots the velocities and steering angles over the trajectory progress. 
\subsubsection{Collision Checking and Assessment of Global Safe Progress}
To assess the global safe progress from the current vehicle state, all states, i.e., discretization points from the generated trajectories are checked for collisions with the list of surrounding obstacles~$O$. When iterating, the safe progress in meters along each trajectory is accumulated until (if at all) a state with a collision is detected. For performing the collision checks, the obstacles are transformed into the vehicle coordinate frame, located at the~\ac{com}. In Fig.~\ref{fig:accProcedureViz}, safe and unsafe states are plotted. In addition, in Fig.~\ref{fig:accTrajTree}, the dimensions of an obstacle and the vehicle, located at the first state for which a collision is detected, are shown. As visualized, collision checks are performed by approximating the rectangular vehicle as an ellipse and checking whether the edge or corner points of the obstacle lie within. Finally, the safe progress values from all trajectories are compared, and the minimum, i.e., the global safe progress~$s_{\textrm{safe}}$ is forwarded to the velocity optimization procedure. \\
Planning trajectories with constant steering rates does not evaluate all possible operator steering actions. Instead, it would be expected that the planner varies the steering rate along the trajectories. However, while this would increase the density of the trajectory tree, it has been found that the resulting global safe progress and the critical curvature profile, described next, are essentially not affected by this. In consequence, it is deemed sufficient to plan with constant steering rates. 
\subsubsection{Computation of Critical Curvature Profile}
\begin{figure}[!htb]%
\begin{subfigure}{\linewidth}
	\includegraphics[width=\textwidth]{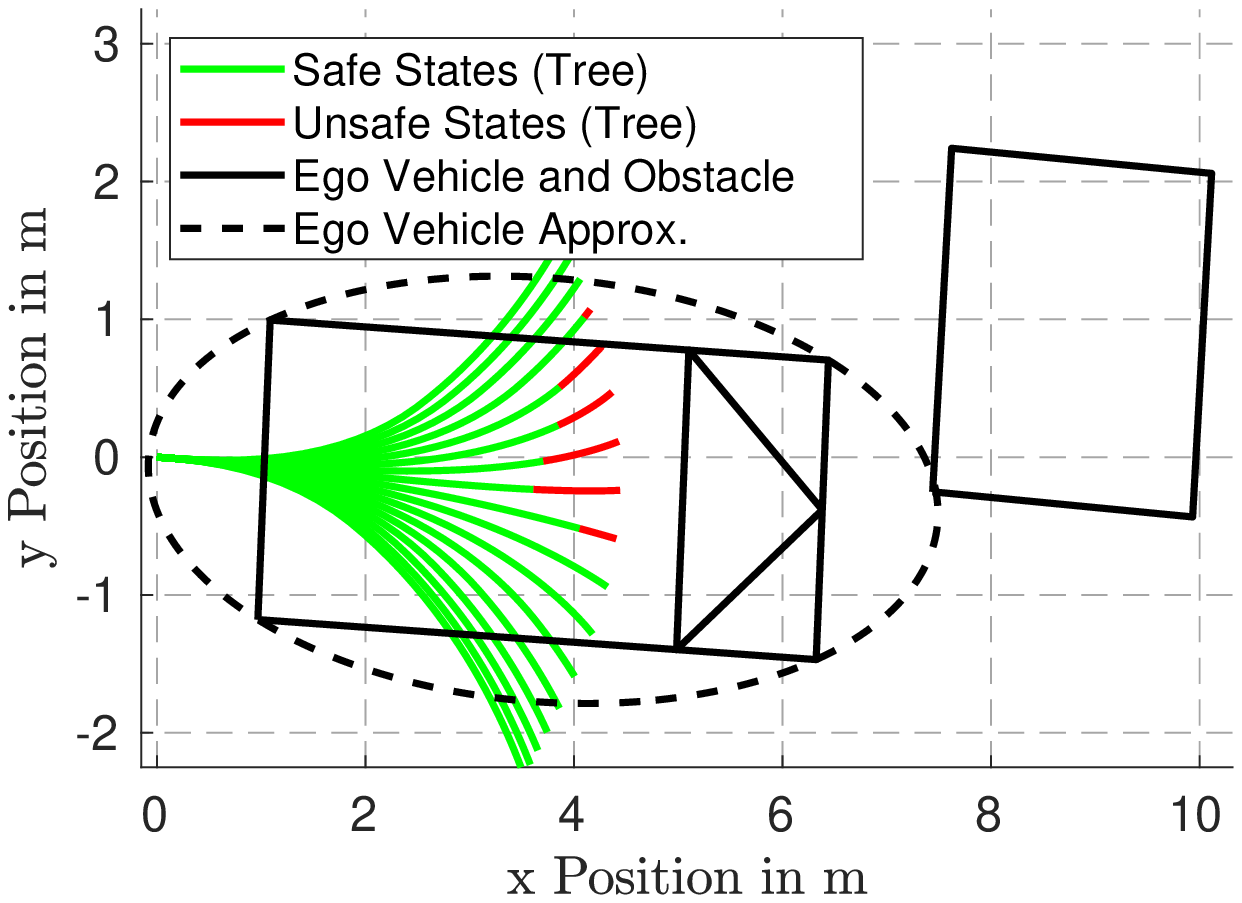}
	\caption{Generated trajectory tree in the $xy$ plane. Safe and unsafe states are depicted. An obstacle and the controlled vehicle, located at the first colliding state with its elliptical approximation, are shown.}
\label{fig:accTrajTree}%
\end{subfigure}
\begin{subfigure}{\linewidth}
	\includegraphics[width=\textwidth]{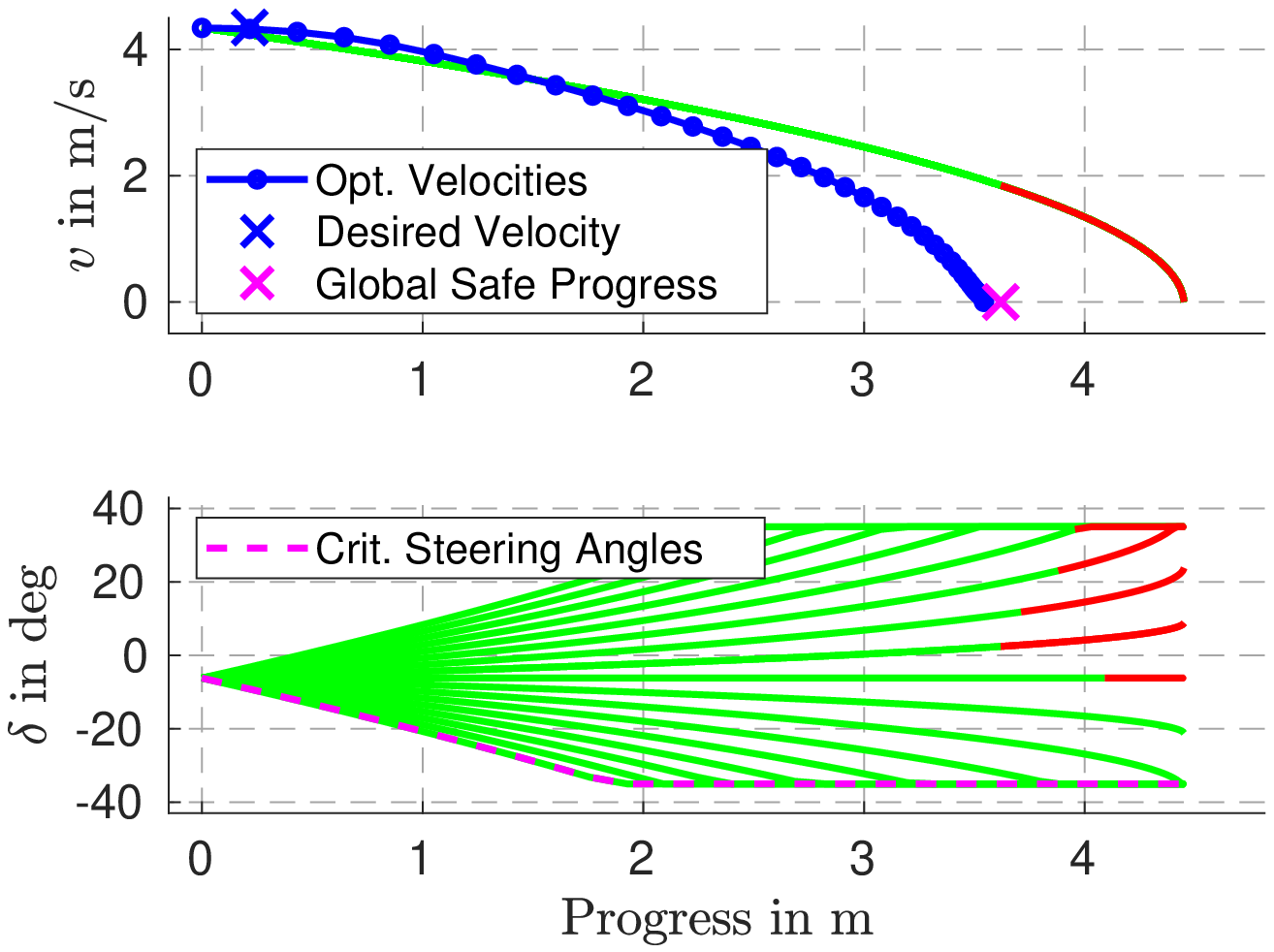}
	\caption{Profiles of velocities (top) and steering angles (bottom) over progress along trajectories in the tree. In the velocity plot, the desired velocity and the global safe progress are marked. Furthermore, the solution of the velocity optimization is shown. In the steering angle plot, the profile of critical steering angles is shown as well.}
	\label{fig:accStatesOverProgress}
\end{subfigure}
\begin{subfigure}{\linewidth}
	\includegraphics[width=\textwidth]{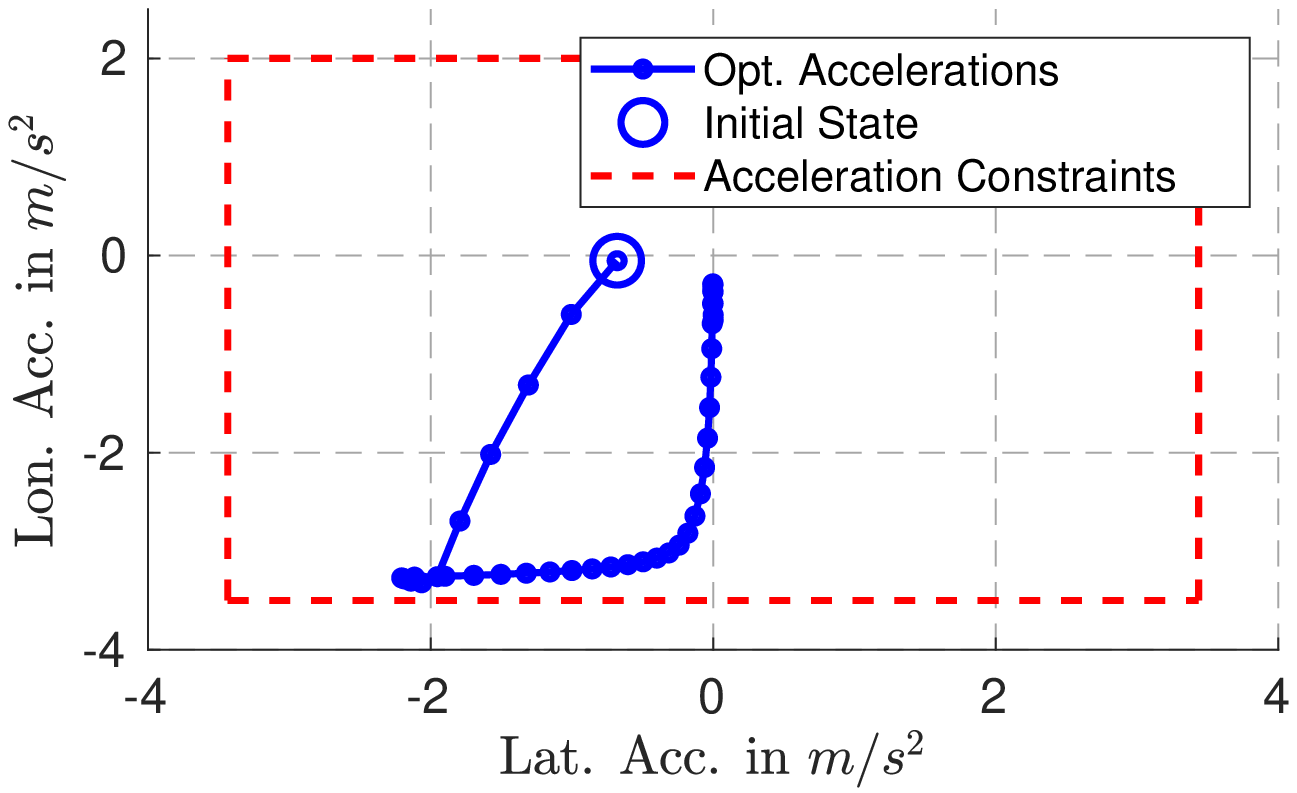}%
    \caption{Acceleration diagram. With the initial state being marked, the acceleration profile, resulting from the velocity optimization, is plotted. In addition, the acceleration constraints are shown.}%
    \label{fig:accGGDiagram}
\end{subfigure}
\caption{Controller computations at one sampling instant.}
\label{fig:accProcedureViz}
\end{figure}

To account for lateral acceleration constraints in the velocity optimization, a critical curvature profile~$\kappa_{\textrm{crit}}^N$ is computed by the trajectory tree planner. This profile stems from the steering angle profile that reaches the maximum steering angle~$\delta_{\textrm{max}}$ the earliest. In short, it is generated by applying the constant steering angle rate~$\dot{\delta}_{\textrm{crit}} = \textrm{sign}(\delta_{\textrm{curr}}) \, \dot{\delta}_{\textrm{max}}$ until the maximum steering angle is reached. Exemplarily, this is shown in the steering angle plot in Fig.~\ref{fig:accStatesOverProgress}. Finally, from each steering angle, the corresponding curvature is computed from the kinematic bicycle model equations. 
%
%
%
%
%
\subsection{Velocity Optimization}
\label{sec:velOptim}
After the trajectory sampling, the velocity profile is optimized in a~\ac{mpc} fashion. On the one hand, the objective is to reach the velocity desired by the human operator. On the other, the velocity profile should brake the vehicle to a standstill, while satisfying safety and comfort constraints. 
\newpage
The optimization problem, also discretized in~$N$ time steps of length~$t_s$, is given by
\begin{subequations}
\begin{equation}
\min \; w_{v,\textrm{des}} \; (v_1 - v_{\textrm{des}})^2 + w_{v,\textrm{term}} \; v_N^2 + J_s
\label{eq:velOptimCost}%
\end{equation}
subject to 
\begin{equation}
s_0 = \SI{0}{\meter} \; , \;  v_0 = v_{\textrm{curr}} \; , \; a_0 = a_{\textrm{curr}},
\label{eq:velOpInitState}%
\end{equation}
\begin{equation}
s_{n+1} = s_{n} + t_s \, v_{n} \; , \;  \; 
v_{n+1} = v_{n} + t_s \, a_{n} \; , \;  \;
a_{n+1} = a_{n} + t_s \, j_{n} \; ,
\label{eq:velOpModelEqs}%
\end{equation}
\begin{equation}
s_{n+1} \leq s_{\textrm{safe}},
\label{eq:velOpProgrConstr}%
\end{equation}
\begin{equation}
-a_{\textrm{lat},\textrm{max}} \leq \kappa_{\textrm{crit},n+1} \, v_{n+1}^2 \leq a_{\textrm{lat},\textrm{max}},
\label{eq:velOpLatAccConstr}%
\end{equation}
\begin{equation}
a_{\textrm{min}} - s_{la,n+1} \leq a_{n+1} \leq a_{\textrm{max}} + s_{ua,n+1},
\label{eq:velOpAccConstr}%
\end{equation}
\begin{equation}
-j_{\textrm{max}} - s_{lj,n} \leq j_n \leq j_{\textrm{max}} + s_{uj,n},
\label{eq:velOpJerkConstr}%
\end{equation}
\begin{equation*}
\textrm{for}~n=0,1,... \, N-1.%
\end{equation*}%
\label{eq:velOptim}%
\end{subequations}
In the cost function~\eqref{eq:velOptimCost}, weighted by~$w_{v,\textrm{des}}$, the primary objective is to reach the desired velocity~$v_{\textrm{des}}$ at the first predicted instant of the horizon~$v_1$. Heavily weighted by~$w_{v,\textrm{term}}$, the terminal velocity~$v_N$ should reach~$\SI{0}{\meter\per\second}$. The system is modelled as a point mass with system states~$[s,v,a]^T$ and jerk~$j$ as control input. The initial state condition and the model equations are given in~\eqref{eq:velOpInitState} and~\eqref{eq:velOpModelEqs}, respectively. Through~\eqref{eq:velOpProgrConstr} and~\eqref{eq:velOpLatAccConstr}, progress and lateral acceleration of the vehicle are constrained to not exceed the global safe progress~$s_{\textrm{safe}}$ and the lateral acceleration limit~$a_{\textrm{lat,max}}$. Being results from the trajectory sampling stage,~$s_{\textrm{safe}}$ and the critical curvature profile~$\kappa^N_{\textrm{crit}}$ are parameters that are updated at each sampling instant of the controller. Finally, through~\eqref{eq:velOpAccConstr} and~\eqref{eq:velOpJerkConstr}, acceleration and jerk are constrained. To cope with noise in the vehicle feedback, and yet achieve the feasibility of the optimization problem, these constraints are made soft. This is achieved by introducing the slack variables~$s_{(\cdot,\cdot)}$, which are penalized in the cost function term~$J_s$. \\
Exemplarily, for the sampling instant visualized in Fig.~\ref{fig:accProcedureViz}, the optimized velocity and acceleration profiles are shown in the velocity plot of Fig.~\ref{fig:accStatesOverProgress} and the acceleration diagram in Fig.~\ref{fig:accGGDiagram}, respectively. Although not all trajectories of the tree are collision-free, the velocity profile can be planned, maintaining the desired velocity and reaching a standstill at the global safe progress. Neither the longitudinal nor the lateral accelerations exceed the imposed constraints. 
\section{\uppercase{Simulation Results}}
\label{sec:simResults}
The proposed \ac{acc} approach has been validated in the software framework for \ac{tod}, introduced in~\cite{Schimpe21oss4tod}. The characteristics of the approach are compared to those of a baseline shared control approach, capable of also overriding the steering actions from the human operator. 
\subsection{Simulation Setup}
\label{sec:simSetup}
\begin{figure}[!t]%
\begin{subfigure}{\linewidth}
    \includegraphics[width=\linewidth]{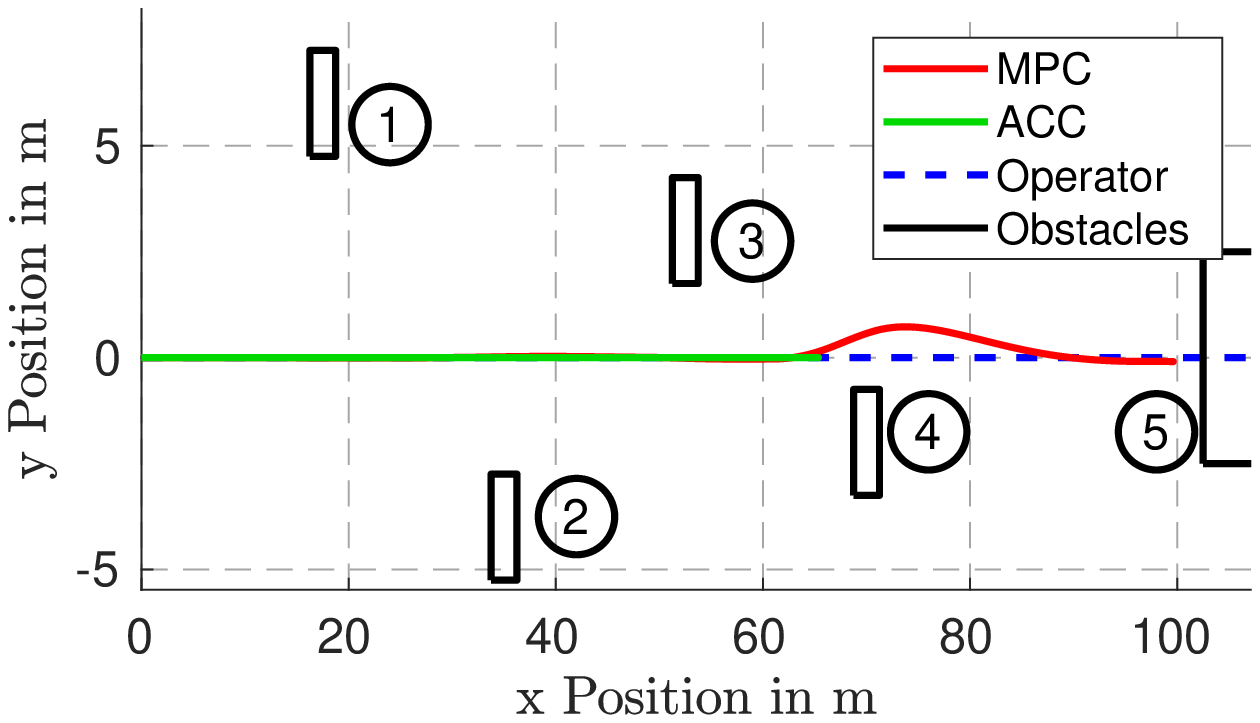}
    \caption{Trajectories in the $xy$ plane for the two control approaches. In addition, the operator reference and obstacles are shown.}
    \label{fig:mpcVSaccXYTrajs}
\end{subfigure}
\begin{subfigure}{\linewidth}
    \includegraphics[width=\linewidth]{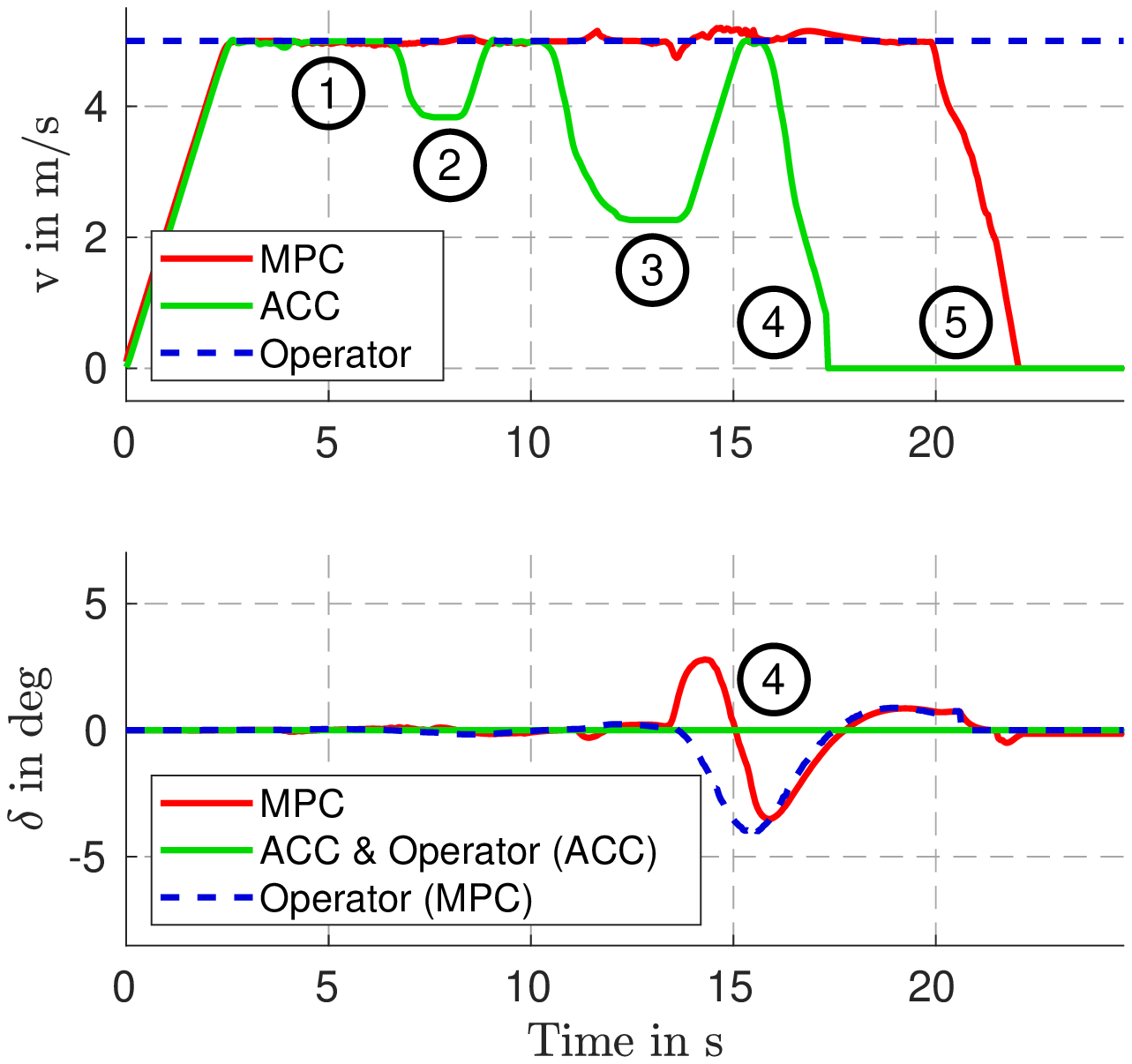}
    \caption{Courses of the actual and desired velocities (top) and steering angles (bottom) over time for the two control approaches.}
    \label{fig:mpcVSaccStates}
\end{subfigure}
\caption{Comparison of \ac{acc} and \ac{mpc} approaches.}
\label{fig:mpcVSacc}
\end{figure}

In the presented simulations, the steering actions from the human operator are simulated through a feedback linearization-based path tracking controller~\cite{Burnett2019buildselfdrivingcar}, following a straight reference path with a constant desired velocity of~\SI{5}{\meter\per\second}. The scenario, visualized in the $xy$ plane in Fig.~\ref{fig:mpcVSaccXYTrajs}, consists of five static obstacles. With decreasing displacement in the $y$ direction, obstacles~\numcircledmod{1}-\numcircledmod{4} are placed alternatingly left and right of the operator reference path. The wider obstacle~\numcircledmod{5} is centered on the reference. \\
The baseline shared control approach is based on \ac{mpc}. Closely following the formulation presented in~\cite{Saparia21ass4tod}, it is capable of velocity and steering intervention. Early in the prediction horizon, the \ac{mpc} aims to track the current operator control command. Obstacle avoidance constraints lead to interventions of the~\ac{mpc} if the operator's control actions would not be collision-free. \\
Both control approaches,~\ac{acc} and~\ac{mpc}, are running at~\SI{20}{\hertz}, a rate easily achieved, given average computation times per sampling instant of less than~\SI{10}{\milli\second}. All prediction horizons, those from the trajectory tree, velocity optimization and \ac{mpc}, are set to~$T_H = \SI{2}{\second}$, discretized in~$N=40$ steps, yielding a sampling time of~$t_s = \SI{50}{\milli\second}$. \\
The optimization problems of the velocity optimization and the~\ac{mpc} are solved using \texttt{acados}~\cite{Verschueren2021acados}, a software package for fast embedded optimization of nonlinear~\ac{mpc} formulations. Internally, \texttt{acados} implements a~\ac{sqp} algorithm, in which QP subproblems are treated by an efficient QP solver. In this work, the QP solver of choice is~\texttt{HPIPM}~\cite{Frison2020hpipm}. 
\subsection{Results}
The performance of the two different shared control approaches in the given scenario is shown in Fig.~\ref{fig:mpcVSacc}. In Fig.~\ref{fig:mpcVSaccXYTrajs}, the driven trajectories are shown in the $xy$ plane. Together with the human operator references, the vehicle velocities and steering angles are plotted over time in Fig.~\ref{fig:mpcVSaccStates}. \\
As shown, the~\ac{acc} tracks the reference velocity, when passing obstacle~\numcircledmod{1}. In proximity to the obstacles~\numcircledmod{2} and~\numcircledmod{3}, as a smaller~$s_{\textrm{safe}}$ is computed, the velocity is reduced to approximately~\SI{4}{\meter\per\second} and~\SI{2}{\meter\per\second}, respectively. Obstacle~\numcircledmod{4} could not be passed without steering. In consequence, at approximately~$x = \SI{65}{\meter}$, the~\ac{acc} brings the vehicle to a standstill. A video of the \ac{acc} simulation is available\footnote{Video of \ac{acc} simulation: \href{https://youtu.be/yFzSiwtUtq4}{https://youtu.be/yFzSiwtUtq4}}. \\
In contrast, the~\ac{mpc} tracks the reference velocity and steering angle consistently when obstacles~\numcircledmod{1} to~\numcircledmod{3} are passed. Obstacle~\numcircledmod{4} is avoided through a minor steering intervention, i.e., deviation from the operator reference. In order to pass obstacle~\numcircledmod{5}, the steering intervention would need to be much greater. Instead, at approximately~$x = \SI{100}{\meter}$, the~\ac{mpc} brings the vehicle to a standstill as well. A video of the \ac{mpc} simulation is available\footnote{Video of \ac{mpc} simulation: \href{https://youtu.be/vz8slCFW140}{https://youtu.be/vz8slCFW140}}. 
\newpage
\section{\uppercase{Experimental Results}}
\label{sec:expResults}
\begin{figure}[!t]%
\includegraphics[width=\linewidth]{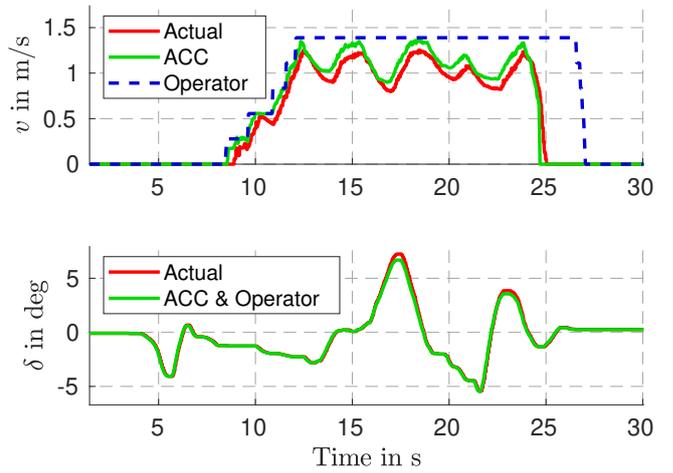}
\caption{Courses of the velocities (top) and steering angles (bottom) desired by the operator, commanded by the ACC and actually executed by the vehicle over time during the teleoperation experiment with the 1:10-scale vehicle testbed.}
\label{fig:expStates}
\end{figure}
\begin{figure*}[!t]
\centering
\begin{subfigure}[!b]{0.48\linewidth}
\includegraphics[width=\linewidth]{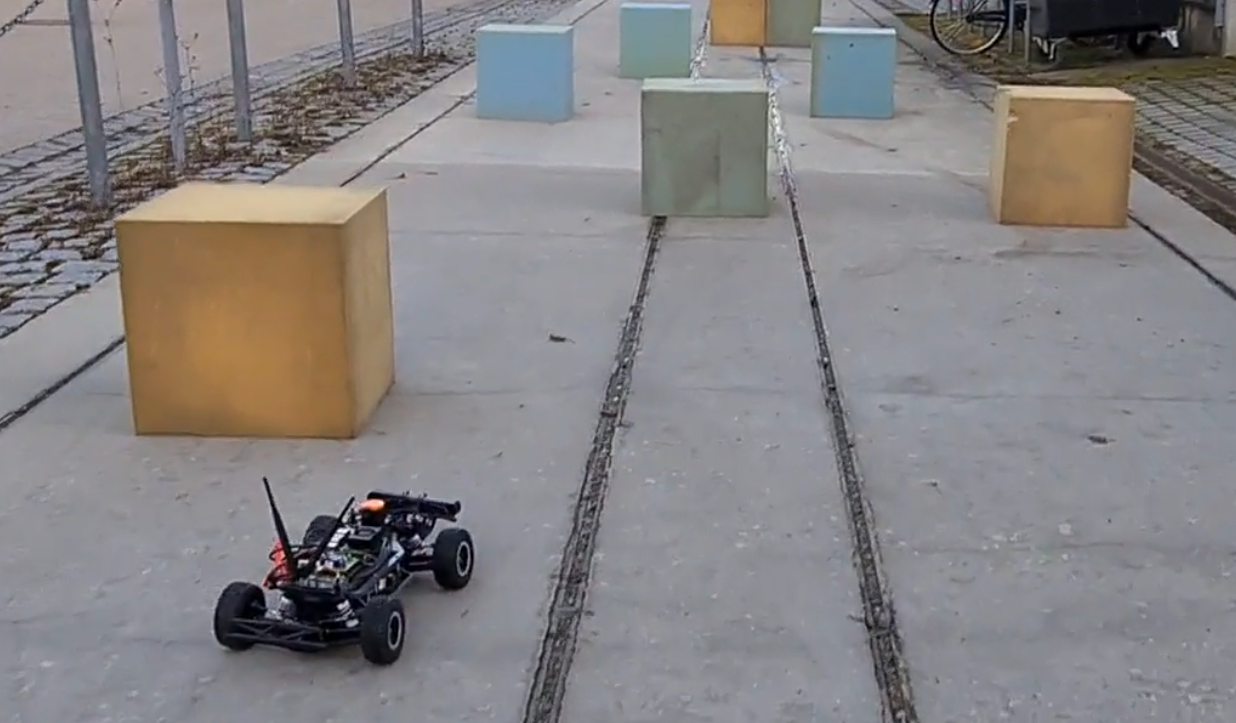}
\end{subfigure}
\hspace{0.01cm}
\centering
\begin{subfigure}[!b]{0.445\linewidth}
\includegraphics[width=\linewidth]{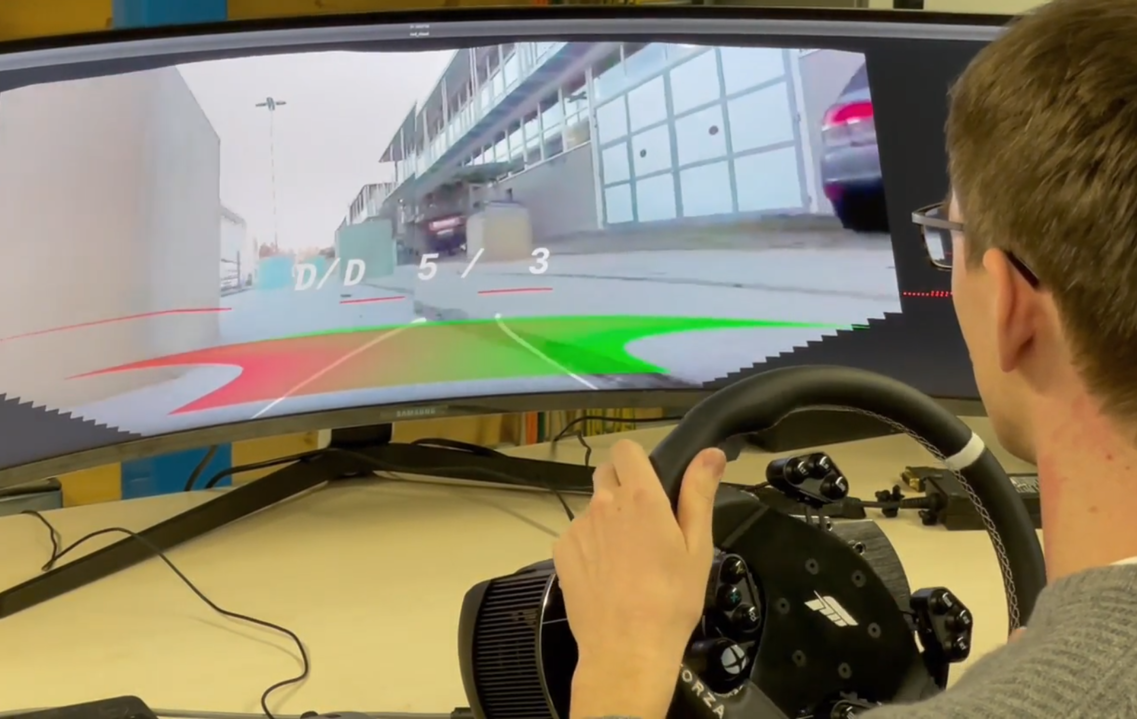}
\end{subfigure}
\caption{Setup of the teleoperation experiment with the 1:10-scale vehicle testbed. The left picture depicts the vehicle and foam cubes used as obstacles. On the right picture, the \ac{vcoc} setup is shown with the operator, the monitor with live videos transmitted from the vehicle, and the steering wheel used to generate the remote control commands.}
\label{fig:expSetup}
\end{figure*}

%
The \ac{acc} approach has also been validated experimentally with a 1:10-scale vehicle testbed~\cite{OKelly2019f1tenth}. The setup of the experiment is shown in Fig.~\ref{fig:expSetup}. The vehicle was teleoperated from the \ac{vcoc} through a course with several obstacles, which were detected by clustering the returns from the onboard 2D lidar sensor. Video streams of the onboard stereo camera from the vehicle as well as control commands from the human operator were transmitted via wireless LAN. \\
The course of the desired and actual velocities as well as steering angles are plotted in Fig.~\ref{fig:expStates}. The controller keeps the vehicle in a safe state, reducing the velocity in proximity to the encountered obstacles. At the time of approximately~\SI{25}{\second}, the operator's steering actions were no longer collision-free. In consequence, the vehicle is brought to a standstill by the \ac{acc}. A video of the experiment is available\footnote{Video of teleoperation experiment: \href{https://youtu.be/dumicX-PZow}{https://youtu.be/dumicX-PZow}}.
\section{\uppercase{Conclusion}}
\label{sec:conclusion}
In this paper, a steering action-aware \ac{acc} approach for teleoperated road vehicles has been proposed. Without the need to override the human operator's steering actions, it was demonstrated that the approach can keep the vehicle in a safe state, i.e., the vehicle can be stopped safely at all times. In simulations, the characteristics of the \ac{acc} approach were compared to those of a baseline shared control approach based on \ac{mpc}, capable of also overriding the operator's steering actions. Finally, the \ac{acc} was validated in a teleoperation experiment with a 1:10-scale vehicle testbed. \\
In future work, the trajectory sampling of the \ac{acc} will be extended to account for dynamic obstacles. Furthermore, the shared control framework will be complemented by visual and haptic feedback in order to establish an advanced bilateral communication channel between the shared controllers and the human operator. The effects on vehicle operability of the approaches will then be assessed in a human subjects study. 
\section*{\uppercase{Acknowledgements}}
Andreas Schimpe developed and implemented the shared control approaches. In fruitful discussions, Domagoj Maj\-storovic contributed to their refinement. The experiments with the 1:10-scale vehicle testbed were supported by Tanay Dwivedi and Florian Sauerbeck. Frank Diermeyer contributed to the essential concept of the research project. 
He revised the manuscript critically for important intellectual content and gave final approval of the version to be published. He agrees with all aspects of the work. 
The research was partially funded by the European Union (EU) under RIA grant No.~825050, and through basic research funds from the Institute for Automotive Technology. 
%
%
\bibliographystyle{IEEEtranBST/IEEEtran}
\bibliography{IEEEtranBST/IEEEabrv,literature}%
\end{document}

%% file: acronyms.tex
\begin{acronym}
\acro{ad}[AD]{automated driving}
\acro{av}[AV]{automated vehicle}
\acroplural{av}[AVs]{automated vehicles}
\acro{com}[CoM]{center of mass}
\acro{hmi}[HMI]{human-machine interface}
\acro{mpc}[MPC]{Model Predictive Control}
\acro{ros}[ROS]{Robot Operating System}
\acro{acc}[ACC]{Adaptive Cruise Control}
\acro{sqp}[SQP]{Sequential Quadratic Programming}
\acro{sa}[SA]{situation awareness}
\acro{tod}[ToD]{Teleoperated Driving}
\acro{ugv}[UGV]{unmanned ground vehicle}
\acroplural{ugv}[UGVs]{unmanned ground vehicles}
\acro{vcoc}[VCoC]{vehicle control center}

\end{acronym}